\documentclass[sigconf]{acmart}

\usepackage{epsfig}
\usepackage{graphicx}
\usepackage{amsmath}
\usepackage{helvet}  
\usepackage{courier}  
\usepackage{url}  
\usepackage{xcolor}
\usepackage{multirow}
\usepackage{subcaption}
\usepackage{array}

\usepackage{times}

\usepackage{balance}

\newcolumntype{P}[1]{>{\centering\arraybackslash}p{#1}}

\def\BibTeX{{\rm B\kern-.05em{\sc i\kern-.025em b}\kern-.08emT\kern-.1667em\lower.7ex\hbox{E}\kern-.125emX}}
    
\copyrightyear{2021}
\acmYear{2021}
\setcopyright{acmlicensed}
\acmConference[ICMR '21]{Proceedings of the 2021 International Conference on Multimedia Retrieval}{August 21--24, 2021}{Taipei, Taiwan}
\acmBooktitle{Proceedings of the 2021 International Conference on Multimedia Retrieval (ICMR '21), August 21--24, 2021, Taipei, Taiwan}
\acmPrice{15.00}
\acmDOI{10.1145/nnnnnnn.nnnnnnn}
\acmISBN{978-x-xxxx-xxxx-x/YY/MM}  

\begin{document}

\fancyhead{}
%
\title{GPT2MVS: Generative Pre-trained Transformer-2 for Multi-modal Video Summarization}

\author{Jia-Hong Huang$^{1^{*}}$, Luka Murn$^{2}$, Marta Mrak$^{2}$, Marcel Worring$^{1}$}
\affiliation{
  \institution{$^{1}$University of Amsterdam, Amsterdam, Netherlands, $^{2}$BBC Research and Development, London, UK  
  \\ $^*$Work done during an internship at BBC Research and Development, London, UK.}
  }
\email{j.huang@uva.nl, luka.murn@bbc.co.uk, Marta.Mrak@bbc.co.uk, m.worring@uva.nl}






%
\begin{abstract}
Traditional video summarization methods generate fixed video representations regardless of user interest. Therefore such methods limit users' expectations in content search and exploration scenarios. Multi-modal video summarization is one of the methods utilized to address this problem. When multi-modal video summarization is used to help video exploration, a text-based query is considered as one of the main drivers of video summary generation, as it is user-defined. Thus, encoding the text-based query and the video effectively are both important for the task of multi-modal video summarization. In this work, a new method is proposed that uses a specialized attention network and contextualized word representations to tackle this task. The proposed model consists of a contextualized video summary controller, multi-modal attention mechanisms, an interactive attention network, and a video summary generator. Based on the evaluation of the existing multi-modal video summarization benchmark, experimental results show that the proposed model is effective with the increase of $+5.88$\% in accuracy and $+4.06$\% increase of F1-score, compared with the state-of-the-art method.
\textbf{\href{https://github.com/Jhhuangkay/GPT2MVS-Generative-Pre-trained-Transformer-2-for-Multi-modal-Video-Summarization}{\scriptsize{https://github.com/Jhhuangkay/GPT2MVS-Generative-Pre-trained-Transformer-2-for-Multi-modal-Video-Summarization}}.}
\end{abstract}

%
%


%

%


%
\maketitle

\begin{figure}[!ht]
\begin{center}
\includegraphics[width=1.0\linewidth]{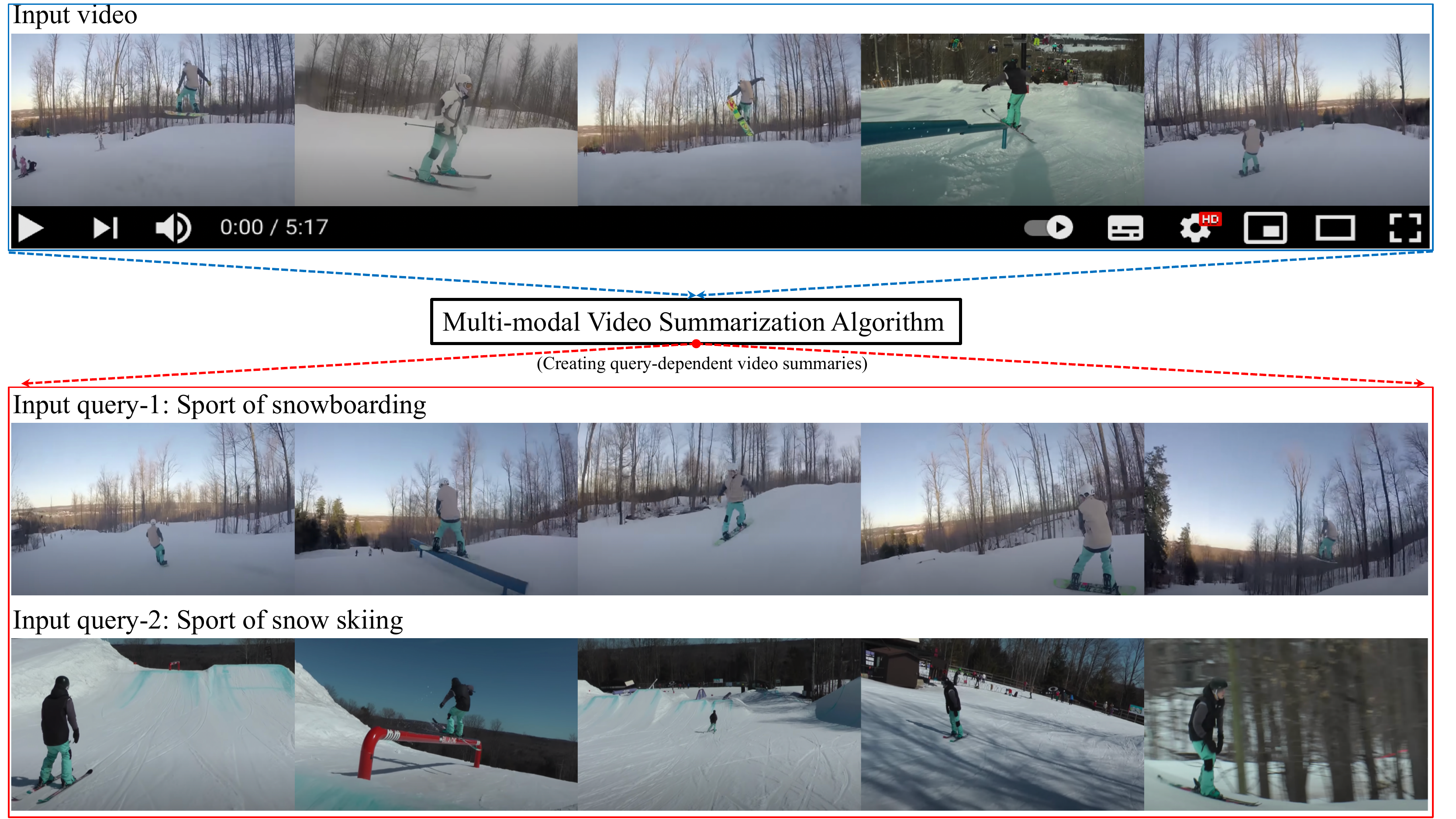}
\end{center}
\vspace{-0.5cm}
   \caption{Multi-modal video summarization. The input video is summarised taking into account text-based queries. ``Input query-1: Sport of snowboarding'' and ``Input query-2: Sport of snow skiing' independently drive the algorithm to generate summaries that contain snowboarding-related content and skiing-related content, respectively.}
\vspace{-0.5cm}
\label{fig:figure1}
\end{figure}

\section{Introduction} 
Video summarization automatically generates a short video clip that summarizes the content of an original, longer video by capturing its important parts \cite{gong2014diverse,zhang2016summary,zhang2019dtr,zhou2018deep}. However, conventional video summarization approaches, such as \cite{ngo2003automatic,song2015tvsum,de2011vsumm,chu2015video,kang2006space,lee2012discovering,gygli2014creating}, only generate a fixed video summary for a given video. Hence, they reduce the effectiveness of video exploration.

Multi-modal video summarization has been proposed as a method to make video exploration more efficient and effective \cite{vasudevan2017query,huang2020query}. The main idea of multi-modal video summarization is to generate video summaries for a given video based on the information provided by the user, i.e., using text-based query to control the video summary, visualized in Figure \ref{fig:figure1}. Traditional video summarization only has one input modality, i.e., video, while an efficient choice for multi-modal video summarization is a text-based query, in addition to video \cite{vasudevan2017query,huang2020query}. Since the text-based query is considered as a controller of the video summary \cite{huang2020query}, effectively encoding the text-based query and capturing the implicit interactions between the query and the video is important. In \cite{huang2020query}, the authors exploit the Bag of Words (BoW) method to encode the query input for multi-modal video summarization. Although BoW has been used with great success on many tasks, such as document classification and language modeling, the authors of \cite{scott1998text,soumya2014text} indicate that BoW suffers from several shortcomings. First, the vocabulary requires careful design because the size of the vocabulary affects the sparsity of the text representation. From the space and time complexity point of view, sparse representations are harder to model. In particular, as the information in such large representation space is sparse, BoW models cannot achieve sufficient effectiveness. Second, since BoW does not encode the order information of a sequence of words, the contextual/semantic meaning likely cannot be captured effectively. Commonly used methods \cite{antol2015vqa,huang2020query,huang2018robustness}, e.g., summation, element-wise multiplication, and concatenation can be used to encode interactive information between the textual and visual features. Although these methods are capable of capturing some interactions between the textual information and visual information, they are still not effective enough and suffer from some information loss \cite{huang2019novel,huang2017robustness,huang2017vqabq,antol2015vqa,huang2020query}.

In this work, a new method is proposed that tackles the aforementioned issues to improve the performance of a multi-modal video summarization model. As stated in \cite{ethayarajh2019contextual}, the commonly used method of static word embeddings, e.g., skip-gram with negative sampling (SGNS) \cite{mikolov2013distributed} or global vectors for word representation (GloVe) \cite{pennington2014glove}, is a better way to encode the text-based query than BoW. However, since SGNS and GloVe generate a single representation for each word, a notable limitation
with static word embeddings is that all senses of a polysemous word must share a single vector \cite{ethayarajh2019contextual}. According to \cite{ethayarajh2019contextual}, the method of contextualized word representations, e.g., Generative Pretrained Transformer-2 (GPT-2), is more effective than static word embeddings. For text-based query encoding, the proposed approach, described in Section 3, exploits the specialized attention network and the contextualized word representations method, i.e., GPT-2, to more effectively encode input text-based query. Also, an attention mechanism with a pre-trained convolutional neural network (CNN) is applied to effectively encode input video. For the interactions between the query and the video, a CNN-based interactive attention module is proposed that can better capture the interactive information.

Typically, the generated video summary by a video summarization algorithm is composed of a set of representative video frames or video fragments \cite{apostolidis2021video}. According to \cite{calic2007efficient,wang2007video,apostolidis2021video}, frame-based video summaries are not restricted by synchronization or timing issues and, therefore, they provide more flexibility in terms of data organization for video exploration purpose. In this work, the proposed model is validated on the frame-based multi-modal video summarization dataset \cite{huang2020query}. 


\vspace{+3pt}
\noindent\textbf{Contributions.}
\begin{itemize}

    \item A new end-to-end deep model for multi-modal video summarization is introduced, based on a specialized attention network and contextualized word representations.

    \item A CNN-based interactive attention network is proposed, in order to better capture the implicit interactive information between the query and the video.

    \item The proposed method is thoroughly validated through experiments on the existing multi-modal video summarization dataset. The experimental results show that the proposed model is effective and achieves state-of-the-art performance, increasing both the accuracy and F1-score.
    
\end{itemize}

\noindent
The rest of the paper is organized as follows: In Section 2, the related work is reviewed. Then, the proposed method is introduced in Section 3. Finally, an evaluation on the effectiveness of the proposed method is conducted in Section 4, followed by a discussion of the experimental results.

\section{Related Work}
In this section, related work in terms of different video summarization methods and word embedding approaches is presented. Two main types of methods of video summarization are discussed, i.e., video summarization with single modality and multi-modal video summarization. Then, word embedding methods are reviewed.

\vspace{+10pt}\noindent\textbf{2.1 Video Summarization with Single Modality}

There are several methods that model the problem of video summarization with single modality, i.e., supervised, weakly supervised, and unsupervised approaches.

\noindent\textbf{Supervised.}
Supervised learning methods for video summarization, e.g., 
\cite{gygli2014creating,gong2014diverse,zhang2016video,zhao2017hierarchical,huang2021deepopht,zhao2018hsa,hu2019silco,zhang2019dtr,ji2019video,ji2020deep}, usually exploit ground truth video summaries, i.e., human expert labeled data, to supervise their models in the training phase. \cite{gygli2014creating} proposed a video summarization method focused on user videos containing a set of interesting events. Their method starts by segmenting a video based on a superframe segmentation, tailored to raw videos. Then, the authors exploit various levels of features to estimate the score of visual interestingness per superframe. A final video summary is generated by selecting a set of superframes in an optimised way. In \cite{gong2014diverse}, the authors model video summarization as a supervised subset selection problem and propose a probabilistic model for selecting a diverse sequential subset, i.e., the sequential determinantal point process (SeqDPP). The SeqDPP is capable of modeling diverse subsets, which is essential for video summarization because it heeds the inherent sequential structures in video data. This overcomes the deficiency of the standard DPP, which treats video frames as randomly permutable elements. \\
An early deep-learning-based method \cite{zhang2016video} considered video summarization as a structured prediction problem and estimated the importance of video frames by modeling their temporal dependency.
The Long Short-Term Memory (LSTM) unit \cite{hochreiter1997long} is used to model the variable-range temporal dependency among frames. Based on the Determinantal Point Process (DPP) \cite{kulesza2012determinantal}, the diversity of visual content of the generated video summary is increased. The authors exploited a multilayer perceptron (MLP) to estimate the importance of video frames. Recurrent Neural Network (RNN) architectures have been used in a hierarchical way to model the temporal structure \cite{zhao2017hierarchical,zhao2018hsa}. This knowledge is utilized to select the video fragments of the summary. To deal with the frame-based video summarization problem, \cite{zhang2019dtr} proposed a dilated temporal relational generative adversarial network (DTR-GAN). The Dilated Temporal Relational (DTR) and LSTM units are combined to estimate temporal dependencies among video frames at different temporal windows. When distinguishing the machine-based video summary from the ground-truth and a randomly-created one, the proposed model learns the summarization task by fooling a trainable discriminator. The authors of \cite{ji2019video} consider video summarization as a sequence-to-sequence learning problem and introduce an LSTM-based encoder-decoder architecture with an intermediate attention layer. This model has later been extended by integrating a semantic preserving embedding network \cite{ji2020deep}.

\noindent\textbf{Weakly Supervised.}
Video summarization has also been considered as a weakly-supervised learning problem \cite{panda2017weakly,ho2018summarizing,cai2018weakly,chen2019weakly}. Similar to unsupervised learning approaches, weakly-supervised methods try to mitigate the need for extensive human-generated ground-truth data. Weakly-supervised methods exploit less-expensive weak labels, e.g., video-level metadata or ground-truth annotations for a small subset of frames, to train models instead of using no ground-truth data. The hypothesis of weakly-supervised learning asserts that weak labels can be used to train video summarization models effectively, even though they are imperfect compared to a full set of human annotations. \\
The authors of \cite{panda2017weakly} were the first to introduce a method adopting an intermediate way between unsupervised and supervised learning for video summarization, i.e., a weakly supervised learning method. They exploited video-level metadata, such as a video title, to define a categorization of videos. Then, multiple videos of a category were leveraged to extract 3D-CNN features and learn a parametric model for categorizing new videos. Finally, the trained model is used to select the video segments that maximize the relevance between the video category and the summary. The authors of \cite{ho2018summarizing} claimed that collecting a large amount of fully-annotated first-person videos with ground-truth annotations is more difficult than the annotated third-person videos. So, they proposed a weakly-supervised model trained on a set of third-person videos with fully annotated highlight scores and a set of first-person videos where only a small portion of them comes with ground-truth annotations. The authors of \cite{cai2018weakly} introduced a weakly-supervised video summarization model that combines the architectures of the Variational AutoEncoder (VAE) \cite{kingma2013auto} and the encoder-decoder with a soft attention mechanism. In the proposed architecture, the goal of VAE is to learn the latent semantics from web videos. The proposed model is trained by a weakly-supervised semantic matching loss to learn video summaries. In \cite{chen2019weakly}, the authors exploited the principles of reinforcement learning to train a video summarization model based on a set of handcrafted rewards and a limited set of human annotations. The proposed method applied a hierarchical key-fragment selection process with the process divided into several sub-tasks. Each task is learned through sparse reinforcement learning and the final video summary is generated based on rewards about its representativeness and diversity.

\noindent\textbf{Unsupervised.}
Given the lack of any ground-truth data for learning video summarization, most of existing unsupervised methods, e.g.,  \cite{zhao2014quasi,chu2015video,panda2017collaborative,yang2018auto,liu2018synthesizing,mahasseni2017unsupervised,rochan2019video,herranz2012scalable,apostolidis2019stepwise,jung2019discriminative,yuan2019cycle,yang2018novel,apostolidis2020unsupervised}, rely on the rule that a representative video summary ought to assist the viewer to infer the original video content. In \cite{zhao2014quasi}, the authors proposed a method that learned a dictionary from the video based on group sparse coding. Then, a video summary is created by combining segments that cannot be reconstructed sparsely based on the dictionary. The authors of \cite{chu2015video} claimed that important visual concepts usually appear repeatedly across videos with the same topic. Therefore, they proposed a maximal biclique finding (MBF) algorithm to find sparsely co-occurring patterns. The video summary is generated based on finding shots that co-occur most frequently across videos. In \cite{panda2017collaborative}, the authors introduced a video summarization approach that is capable of simultaneously capturing the generalities identified from a set of given videos and the particularities arising in a given video. \\
The authors of \cite{mahasseni2017unsupervised} proposed a video summarization method that combines a trainable discriminator and an LSTM-based key-frame selector with a VAE. The proposed model learns to generate video summaries based on an adversarial learning process. The process aims to minimize the distance between the original video and the summary-based reconstructed version of the original video. Based on the network proposed by \cite{mahasseni2017unsupervised}, a stepwise label-based method for training the adversarial part of the network has been suggested in order to improve the model performance \cite{apostolidis2019stepwise}. Similarly, a method based on a VAE-GAN architecture has been proposed \cite{jung2019discriminative}. The model is extended with a chunk and stride network (CSNet). The authors of \cite{rochan2019video} introduced a new formulation to perform video summarization from unpaired data. The goal of the proposed method is to learn a mapping such that the distribution of the generated video summary is similar to the distribution of the set of video summaries with the help of an adversarial objective. A diversity constraint is also enforced on the mapping to ensure the generated video summaries are visually diverse enough. The authors of \cite{yuan2019cycle} proposed a method to maximize the mutual information between the video and video summary based on a trainable couple of discriminators and a cycle-consistent adversarial learning objective. A variation of \cite{apostolidis2019stepwise} has been proposed in \cite{apostolidis2020unsupervised}, where the VAE is replaced with a deterministic attention auto-encoder for learning an attention-driven reconstruction of the original video. This subsequently improves the process of key-fragment selection.

Typically, supervised methods are capable of learning useful cues, which are hard to capture from ground truth summaries with hand-crafted heuristics. Therefore, supervised approaches usually outperform the weakly supervised and unsupervised models. In this work, video summarization is modeled as a supervised learning task.

\vspace{+10pt}\noindent\textbf{2.2 Multi-modal Video Summarization}

Instead of only considering the visual input, a number of works have investigated the potential of using some additional modality, such as video captions, viewers’ comments, or any other available contextual data, for learning video summarization \cite{li2017extracting,vasudevan2017query,sanabria2019deep,song2016category,zhou2018video,huang2021contextualized,lei2018action,huang2019assessing,otani2016video,huang2021deep,yuan2017video,wei2018video,huang2020query}. The authors of \cite{li2017extracting} introduced a multi-modal video summarization method for key-frame extraction from first-person videos. The authors of \cite{sanabria2019deep} proposed a multi-modal deep-learning-based approach to summarize videos of soccer games. Under the context of \cite{song2016category}, a method is introduced that learns the category-driven video summary by rewarding the preservation of the core parts which are found in video summaries from the same category \cite{zhou2018video}. Similarly, the authors of \cite{lei2018action} propose to train action classifiers with video-level annotations for action-driven video fragmentation and labeling. Then, a fixed number of key-frames is extracted and reinforcement learning is applied to select the ones with the highest accuracy of categorization to perform category-driven video summarization. In \cite{otani2016video,yuan2017video}, the authors defined a video summary by maximizing its relevance with the available video metadata, after projecting the textual and visual information in a common latent space. In \cite{wei2018video}, a semantic-based video fragment selection and a visual-to-text mapping is applied based on the relevance between the original and the automatically-generated video descriptions, with the help of semantic attended networks.

Existing multi-modal video summarization approaches exploit static word embeddings \cite{mikolov2013distributed,pennington2014glove} to encode textual input. However, \cite{ethayarajh2019contextual} has shown that static word embedding is not effective enough compared with contextualized word representations. In this work, a new multi-modal video summarization method is introduced, based on the contextualized word representations, to make the video exploration more efficient and effective.

\vspace{+10pt}\noindent\textbf{2.3 Word Embeddings}

According to \cite{ethayarajh2019contextual}, existing word embedding methods are categorized into two categories, i.e., static word embeddings. and contextualized word representations.

\noindent\textbf{Static Word Embeddings.} 
GloVe \cite{pennington2014glove} and Skip-gram with negative sampling (SGNS) \cite{mikolov2013distributed} are among the best-known models for the generation of static word embeddings. In practice, although these models learn word embeddings iteratively, it has been proven that both models implicitly factorize a word-context matrix containing a co-occurrence statistic \cite{levy2014linguistic,levy2014neural}. A limitation of the static word embeddings method is that all meanings of a polysemous word must share a single vector because a single representation for each word is created.

\noindent\textbf{Contextualized Word Representations.} 
To tackle the aforementioned issue with static word embeddings, recent works have been proposed that create context-sensitive word representations \cite{peters2018deep,devlin2018bert,radford2019language}. The deep neural language models proposed by \cite{devlin2018bert,peters2018deep,radford2019language} are fine-tuned to create deep learning based models for a wide range of downstream natural language processing tasks. Since the internal representations of words of the above methods are a function of the entire input query sentence, the representations are called contextualized word representations. The authors of \cite{liu2019linguistic} proposed a successful method which suggests that these representations capture task-agnostic and highly transferable properties of language. The method proposed in \cite{peters2018deep} generated contextualized representations of every token by concatenating the internal states of a 2-layer bi-LSTM which is trained on a bidirectional language modeling task. The approaches proposed by \cite{radford2019language,devlin2018bert} are uni-directional and bi-directional transformer-based language models, respectively. In \cite{ethayarajh2019contextual}, the author has shown the method of static word embeddings is less effective than the contextualized word representations. Therefore, in this work, a new approach is proposed that uses a specialized attention network and the contextualized word representations method for query encoding.

\begin{figure*}[!ht]
  \includegraphics[width=\textwidth]{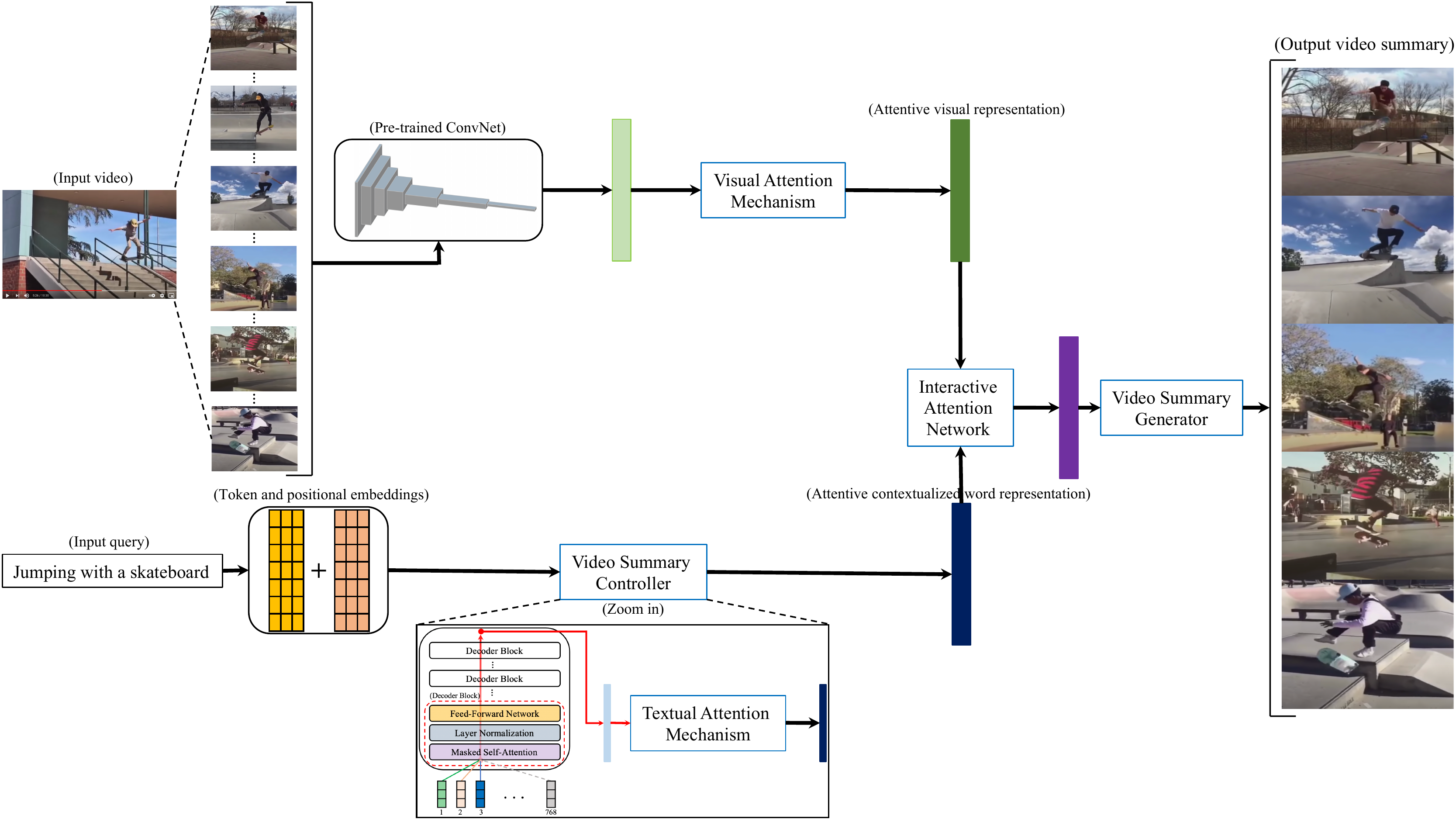}
    \vspace{-0.7cm}
  \caption{
  Flowchart of proposed multi-modal video summarization method. A pre-trained CNN is used to extract features from the visual input to enable the ``Visual Attention Mechanism'' to generate the attentive visual representation (dark green). From an input text-based query, the ``Token and positional embedding'' generates the input to the ``Video Summary Controller''. The ``Textual Attention Mechanism'' generates the attentive contextualized word representation (dark blue). The ``Interactive Attention Network'' takes the attentive visual representation and the attentive contextualized word representation as inputs to generate an informative feature vector (purple). The informative feature vector is the input of the ``Video Summary Generator'' which creates the query-dependent video summary. Please refer to the \textit{Methodology} section for more details.}
  \label{fig:figure2}
  \vspace{-0.1cm}
\end{figure*}



\section{Methodology}

\noindent\textbf{Overview.}
In this section, a novel multi-modal video summarization method is described in detail. The proposed method is composed of a contextualized video summary controller, a textual attention mechanism, a visual attention mechanism, an interactive attention network, and a video summary generator. The flowchart of the proposed method is presented in Figure \ref{fig:figure2}. A pre-trained CNN, e.g., ResNet \cite{he2016deep}, is used to extract features from the visual input and the ``Visual Attention Mechanism' is exploited to generate the attentive visual representation indicated in dark green. An input text-based query, e.g., ``Jumping with a skateboard'', is sent to ``Token and positional embedding'' for generating the input of the ``Video Summary Controller'' which is composed of a stack of decoder blocks and a ``Textual Attention Mechanism''. Each decoder block consists of masked self-attention, layer normalization, and a feed-forward network, indicated as the red dashed line box, and the $768$ color-coded brick-stacked vectors serving as the input of the summary controller. Note that the masked self-attention is considered as a function of $Q$, $K,$ and $V$, i.e., $MaskAtten(Q,K,V)$ in Equation \ref{eq:attention}. We exploit the ``Textual Attention Mechanism'' with the output from the last decoder block to generate the attentive contextualized word representation colored by dark blue, i.e., the output of the summary controller. The proposed ``Interactive Attention Network'' takes the attentive visual representation and the attentive contextualized word representation as inputs to generate an informative feature vector shown in purple. The informative feature vector is the input of the ``Video Summary Generator''. The job of the summary generator is to output the query-dependent video summary.


\vspace{+10pt}\noindent\textbf{3.1 Contextualized Video Summary Controller}

The transformer architecture \cite{vaswani2017attention} has been firmly established as one of the state-of-the-art methods in machine translation and language modeling. It is mainly composed of a transformer-encoder and a transformer-decoder. The encoder and decoder are stacks of multiple basic transformer blocks. Inspired by the transformer-decoder structure, i.e., GPT-2, in terms of its masked self-attention and parallelization, 
these characteristics are employed to develop the contextualized video summary controller for the text-based query embedding.
The summary controller is described in detail as follows. For an input token $k_n$, its embedding $x_n$ is defined as: 

\begin{equation} 
    x_n = W_e*k_n+P_{k_n}, n \in \{0,...,N-1\}, \\
	\label{eq:text_embed1}
\end{equation} 
where $W_e \in \mathbb{R}^{E_s \times V_s}$ is the token embedding matrix with the word embedding size $E_s$ and the vocabulary size $V_s$, $P_{k_n}$ is the positional encoding \cite{vaswani2017attention} of $k_n$, and $N$ denotes the number of input tokens. $n$ is a non-negative integer from $0$ to $N-1$. The subscript $e$ and the subscript $s$ denote ``embedding'' and ``size'', respectively.



For the representation of the current word $Q$, it is generated by one linear layer which is
defined as: 

\begin{equation}
    Q = W_q*x_n+b_q,
	\label{eq:text_embed2}
\end{equation}
where $W_q \in \mathbb{R}^{H_s \times E_s}$ and $b_q$ are learnable parameters of the linear layer.  $H_s$ is the output size of the linear layer and the subscript $q$ denote ``query'' \cite{vaswani2017attention}.


For the key vector $K$ \cite{vaswani2017attention}, it is calculated by other linear layer which is
defined as: 

\begin{equation}
    K = W_k*x_n+b_k,
	\label{eq:text_embed3}
\end{equation}
where $W_k \in \mathbb{R}^{H_s \times E_s}$ and $b_k$ are learnable parameters of the linear layer. The subscript $k$ denote ``key'' \cite{vaswani2017attention}.


For the value vector $V$ \cite{vaswani2017attention}, it is generated by another linear layer which is
defined as: 

\begin{equation}
    V = W_v*x_n+b_v,
	\label{eq:text_embed4}
\end{equation}
where  $W_v \in \mathbb{R}^{H_s \times E_s}$ and $b_v$ are learnable parameters of the linear layer. The subscript $v$ denote ``value'' \cite{vaswani2017attention}.


After $Q$, $K$, and $V$ are calculated, the masked self-attention $Z$ is generated as Equation-(\ref{eq:attention}).

\begin{equation}
    Z = MaskAtten(Q,K,V) = softmax(m(\frac{QK^T}{\sqrt{d_k}}))V,
	\label{eq:attention}
\end{equation}
where $m(\cdot)$ is a masked self-attention function and $d_k$ denotes a scaling factor \cite{vaswani2017attention}. $Z$ is defined as a function of $Q$, $K$, and $V$.

Then, the layer normalization is calculated as Equation-(\ref{eq:layernorm}).

\begin{equation}
    Z_{Norm} = LayerNorm(Z),
	\label{eq:layernorm}
\end{equation}
where $LayerNorm(\cdot)$ is a function indicating layer normalization.

Through Equations-(1-6) proposed contextualized representation $F$ of the text-based query is derived as:

\begin{equation}
    F = FFN(Z_{Norm}) = \sigma(W_1Z_{Norm}+b_1)W_2+b_2,
	\label{eq:ffn}
\end{equation}
where $FFN(\cdot)$ is a position-wise feed-forward network (FFN), $\sigma$ is an activation function. $W_{1}$, $W_{2}$, $b_{1}$, and $b_{2}$ are learnable parameters of the FFN.

\vspace{+10pt}\noindent\textbf{3.2 Multi-modal Attentions}

To have even better textual and visual representations, a textual attention mechanism is proposed to reinforce the contextualized representation $F$, and a visual attention mechanism to reinforce the visual features extracted by the CNN.

\noindent\textbf{Textual Attention Mechanism.}
The proposed textual attention mechanism is defined as a textual attention function $TextAtten(\cdot)$, referring to Equation-(\ref{eq:text-attention}). The function takes the result of FFN, i.e., $F$ from Equation-(\ref{eq:ffn}), as input and calculates the attention and textual representation in an element-wise way, i.e., Hadamard textual attention.


\begin{equation}
    Z_{ta} = TextAtten(F),
	\label{eq:text-attention}
\end{equation}
where the subscript $ta$ denotes ``textual attention''.

\noindent\textbf{Visual Attention Mechanism.}
The proposed visual attention mechanism is defined as a visual attention function $VisualAtten(\cdot)$, referring to Equation-(\ref{eq:visual-attention}). The function takes visual representation $\phi(I)$, extracted by the CNN, as input and calculates the attention and visual representation in an element-wise way, i.e., Hadamard visual attention.


\begin{equation}
    Z_{va} = VisualAtten(\phi(I)),
	\label{eq:visual-attention}
\end{equation}
where the subscript $va$ denotes ``visual attention''.


\vspace{+10pt}\noindent\textbf{3.3 Interactive Attention Network}

Interactive information between the query and the video is crucial in multi-modal video summarization. An interactive attention network is proposed to more effectively capture this interaction between the query and the video. In Equation-(\ref{eq:inter-attention}), $InterAtten(\cdot)$ denotes an interactive attention network. The network performs one by one convolution, i.e., convolutional attention.

\begin{equation}
    Z_{ia} = InterAtten(Z_{ta} \odot Z_{va}),
	\label{eq:inter-attention}
\end{equation}
where $Z_{ta}$ denotes textual attention, $Z_{va}$ denotes visual attention, and $\odot$ denotes Hadamard product.

\vspace{+10pt}\noindent\textbf{3.4 Loss Function}

In \cite{huang2020query}, the authors consider the task of multi-modal video summarization as a classification problem and take the commonly used classification loss, i.e., cross-entropy loss, as their loss function. In this work, since the multi-modal video summarization problem is modeled as a classification task and validated on the same dataset used in \cite{huang2020query}, the cross-entropy loss function, referring to Equation \ref{eq:loss}, is also adopted to build the proposed model.

\begin{equation}
    Loss(x, class) = -x[class]+ln(\sum_{j}exp(x[j])),
    \label{eq:loss}
\end{equation}
where $x$ denotes the prediction, $class$ indicates the ground truth class, and $j$ denotes the index for iteration \cite{NEURIPS2019_9015,huang2020query}.

\vspace{+10pt}\noindent\textbf{3.5 Video Summary Generator}

The goal of the video summary generator is to create video summaries based on the effective vector representation of the text-based query and video, i.e., the result from Equation-(\ref{eq:inter-attention}). The proposed summary generator exploits the fully-connected linear layer to generate a frame-based score vector for a given query-video pair. Then, it outputs the final video summary based on the vector. Please refer to Figure \ref{fig:figure2} for the entire video summary generation procedure. 


\begin{table*}[!ht]
    \caption{Performance evaluation of the proposed method for different dimensions of contextualized word representations. The best word embedding dimension (numbers in bold) is selected experimentally for each model's accuracy \cite{huang2020query}. Note the default output word embedding dimensions of each model: GPT-2$=768$, GPT-2-M$=1024$, GPT-2-L$=1280$, GPT-2-XL$=1600$.}
    \vspace{-0.4cm}
\begin{center}
\scalebox{1.0}{
    \begin{tabular}{|P{5.0cm}|P{2.0cm}|P{2.cm}|P{2.0cm}|P{2.0cm}|}
    \hline
    \textbf{Word Embedding Dimension} & \textbf{GPT-2} & \textbf{GPT-2-M} & \textbf{GPT-2-L} & \textbf{GPT-2-XL}\\ \hline
    $10$ dimensions   & 0.7551  & 0.7399  & 0.7506  & 0.7547\\ \hline
    $50$ dimensions   & 0.7541  & 0.7194  & 0.7509  & 0.7419\\ \hline
    $100$ dimensions  & 0.7510  & 0.7527  & 0.7428  & 0.7342\\ \hline
    $150$ dimensions  & 0.7447  & 0.7484  & 0.7392  & \textbf{0.7662}\\ \hline
    $200$ dimensions  & 0.7383  & 0.7447  & 0.7035  & 0.7468\\ \hline
    $250$ dimensions  & 0.7355  & 0.7465  & 0.7158  & 0.7447\\ \hline
    $300$ dimensions  & \textbf{0.7566}  & 0.7452  & 0.7552  & 0.7552\\ \hline
    $350$ dimensions  & 0.7543  & \textbf{0.7558}  & 0.7473  & 0.7244\\ \hline
    $400$ dimensions  & 0.7318  & 0.7326  & \textbf{0.7583}  & 0.6922\\ \hline
    $450$ dimensions  & 0.7381  & 0.7470  & 0.7407  & 0.7435\\ \hline
    $500$ dimensions  & 0.7436  & 0.7501  & 0.7451  & 0.7451\\ \hline
    Default output dimensions  & 0.7375  & 0.7411  & 0.7460  & 0.7469\\ \hline

    \end{tabular}}

    \vspace{-0.1cm}
\label{table:table1}
\end{center}
\end{table*}


\begin{table*}[!ht]
\caption{Ablation study of various attentions using $F_{1}$-score \cite{hripcsak2005agreement,gygli2014creating,song2015tvsum} to quantify the model performance. ``w/o'' denotes models without a specific type of attention, and ``w/'' denotes models with a specific type of attention. According to $F_{1}$-scores, the proposed attentions are effective for all tested models.}
\vspace{-0.4cm}
\centering
\scalebox{1.0}{
\begin{tabular}{|c|c|c|c|c|c|}
\hline
\multicolumn{2}{|c|}{\textbf{Attention Type}}   & \textbf{GPT-2} (300-dim) & \textbf{GPT-2-M} (350-dim) & \textbf{GPT-2-L} (400-dim) & \textbf{GPT-2-XL} (150-dim) \\ \hline
\multirow{2}{*}{Visual Attention}           & w/o  & 0.4905  & 0.5199  & 0.5225  & 0.5140  \\ \cline{2-6}

                                    & w/  & \textbf{0.5200}   & \textbf{0.5277}  & \textbf{0.5247}  & \textbf{0.5340}   \\ \hline

\multirow{2}{*}{Textual Attention}              & w/o  & 0.4905  & 0.5199  & 0.5225  & 0.5140   \\ \cline{2-6}

                                    & w/  & \textbf{0.5183}   & \textbf{0.5334}   & \textbf{0.5266}  & \textbf{0.5260}   \\ \hline

\multirow{2}{*}{Visual-Textual Attention}     & w/o  & 0.4905  & 0.5199  & 0.5225  & 0.5140    \\ \cline{2-6}

                                    & w/  & \textbf{0.5247} & \textbf{0.5357}   & \textbf{0.5319}   & \textbf{0.5363}    \\ \hline

\multirow{2}{*}{Interactive Attention}        & w/o  & 0.4905  & 0.5199  & 0.5225  & 0.5140    \\ \cline{2-6}

                                    & w/  & \textbf{0.5040}  & \textbf{0.5327}   & \textbf{0.5275}   & \textbf{0.5389}   \\ \hline

\multirow{2}{*}{Interactive-Visual-Textual Attention}       & w/o  & 0.4905  & 0.5199  & 0.5225  & 0.5140    \\ \cline{2-6}

                                    & w/  & \textbf{0.5410}  & \textbf{0.5484}   & \textbf{0.5473}   & \textbf{0.5420}   \\ \hline

\end{tabular}}
\label{table:table2}
    \vspace{-0.1cm}
\end{table*}

\begin{table*}[!ht]
    \caption{Comparison with the state-of-the-art \textbf{QueryVS }\cite{huang2020query}, based on the metric of accuracy \cite{huang2020query} and $F_{1}$-score \cite{hripcsak2005agreement}. Proposed method outperforms the model in \cite{huang2020query} by $+5.88$\% in accuracy and $+4.06$\% in F1-score.}
    \vspace{-0.5cm}
\begin{center}
\scalebox{1.0}{
    \begin{tabular}{|P{2.6cm}|P{2.7cm}|P{2.7cm}|P{2.7cm}|P{2.7cm}|P{1.9cm}|}
    \hline
    \textbf{Evaluation Metric} & \textbf{GPT-2} (300-dim)& \textbf{GPT-2-M} (350-dim)& \textbf{GPT-2-L} (400-dim)& \textbf{GPT-2-XL} (150-dim) & \textbf{QueryVS }\cite{huang2020query}\\ \hline

    Accuracy \cite{huang2020query}  & 0.7424  & \textbf{0.7625}  & 0.7493  & 0.7510 & 0.7037\\ \hline

    $F_{1}$-score \cite{hripcsak2005agreement}  & 0.5410  & \textbf{0.5484}  & 0.5473  & 0.5420 & 0.5078\\ \hline

    \end{tabular}}
    \label{table:table3}
    \vspace{-0.1cm}
\end{center}
\end{table*}

\section{Experiments and Analysis}

\noindent\textbf{Overview.}
In this section, the experimental setup is described in detail and the proposed multi-modal video summarization model is validated on the existing multi-modal video summarization dataset \cite{huang2020query}. Then, the effectiveness of the proposed various attention-based modules and contextualized word representations is analyzed. Finally, randomly selected qualitative results are displayed.

\vspace{+10pt}\noindent\textbf{4.1 Dataset Preparation and Evaluation Metrics}

\noindent\textbf{Dataset.}
In the experiments undertaken for the purposes of this work, the multi-modal video summarization dataset proposed by \cite{huang2020query} is exploited to validate the introduced model. Their proposed dataset consists of 190 videos, with a duration of two to three minutes for each video, and every video is retrieved based on a given text-based query. The authors separate the entire dataset into splits of 60\%/20\%/20\%, i.e., 114/38/38, for training/validation/testing, respectively. To automatically evaluate multi-modal video summarization methods, annotations from human experts are necessary. So, the authors sample all of the 190 videos at one frame per second (fps) and then exploit Amazon Mechanical Turk (AMT) to annotate every frame with its level of relevance with respect to the given text-based query. The distribution of relevance level annotations is ``Very Good'': 18.65\%, ``Good'': 55.33\%, ``Not good'': 13.03\% and ``Bad'': 12.99\%. The authors \cite{huang2020query} map relevance level annotations, ``Very Good'' to 3, ``Good'' to 2, ``Not Good'' to 1, and ``Bad'' to 0. According to \cite{huang2020query}, a single ground truth relevance level label is created for each query-video pair by merging the corresponding relevance level annotations from AMT workers. Note that, in \cite{huang2020query}'s dataset, the maximum number of words of a query is 8. Additionally, the authors propose to, based on the majority vote rule, evaluate the model performance for a relevance score prediction. That is to say, a predicted relevance level is considered correct when the predicted relevance score is the same as the majority of human annotators' provided score, meaning accuracy is used as the evaluation metric.

\noindent\textbf{Evaluation Metrics.}
In \cite{huang2020query}, the authors use accuracy, based on the predicted and ground truth frame-based scores, to evaluate the performance of their model. Since experiments in this work are based on the dataset proposed by \cite{huang2020query}, the same metric of accuracy is used to quantify the model performance. Also, motivated by \cite{hripcsak2005agreement,gygli2014creating,song2015tvsum}, $F_{\beta}$-score with the hyperparameter $\beta=1$, referring to Equation-(\ref{eq:f1-score}), is exploited to assess the performance of the proposed model by measuring the agreement between the predicted and gold standard scores provided by the crowd. 

\begin{equation}
    F_{\beta}=\frac{1}{N}\sum_{i=1}^{N}\frac{(1+\beta ^{2})\times p_{i}\times r_{i}}{(\beta ^{2}\times p_{i})+r_{i}},
    \label{eq:f1-score}
\end{equation}
where $p_{i}$ denotes $i$-th precision, $r_{i}$ denotes $i$-th recall, $N$ denotes number of $(p_{i}, r_{i})$ pairs, ``$\times$'' denotes scalar product, and $\beta$ is used to balance the relative importance between precision and recall. 

\vspace{+10pt}\noindent\textbf{4.2 Experimental Setup}

The technique of pre-training a CNN on ImageNet \cite{deng2009imagenet} and exploiting it to perform vision-related tasks, i.e., as a visual feature extractor, has been widely used because of its effectiveness. In this work, ResNet \cite{he2016deep} pre-trained on ImageNet is adopted to extract 199 frame-based features for each video, and the feature used is located in the visual layer one layer below the classification layer. Note that the video lengths in \cite{huang2020query}'s dataset are varied, so the number of frames for each video, extracted with fps$=1$ by the authors, is different. The maximum number of frames of a video is $199$ in their proposed dataset. The same video preprocessing approach is followed to initiate these experiments, i.e.,  frame-repeating \cite{huang2020query}, to make all the videos have the same length of $199$. The input frame size of the CNN is 224 by 224 with red, green, and blue channels. Each image channel is normalized by mean $=(0.4280, 0.4106, 0.3589)$ and standard deviation $=(0.2737, 0.2631, 0.2601)$.
Since the implementation is based on the transformer-decoder \cite{vaswani2017attention}, i.e., GPT-2 \cite{radford2019language}, to develop the contextualized video summary controller for the text-based query embedding, using the pre-trained weights of GPT-2 to initialize the summary controller is helpful. According to \cite{radford2019language}, GPT-2 has been pre-trained on a large corpus, with vocabulary size $=50,257$. 
In this work, PyTorch is used for the implementation and to train models with $10$ epochs, $1e-4$ learning rate, and Adam optimizer \cite{kingma2014adam}. For the parameters of the Adam optimizer, $\beta_{1}=0.9$ and $\beta_{2}=0.999$ are the coefficients used for computing moving averages of gradient and its square. The term added to the denominator to improve numerical stability is $\epsilon=1e-8$.

\vspace{+10pt}\noindent\textbf{4.3 Effectiveness Analysis of Various Attentions}

Since the word embedding size/dimension affects the training efficiency and model performance, several experiments are conducted with different word embedding dimensions to analyze the proposed method, referring to Table \ref{table:table1}, Table \ref{table:table2}, and Table \ref{table:table3}. Then, the model with the best word embedding dimension and performance is selected to conduct the ablation study based on the proposed various attentions. 

\noindent\textbf{Textual Attention.}
The ablation study of the textual attention mechanism is presented in Table \ref{table:table2}. The results show that the textual attention is effective. It helps improve the performances of models, i.e., $+2.78$\% for GPT-2, $+1.35$\% for GPT-2-M, $+0.41$\% for GPT-2-L, and $+1.2$\% for GPT-2-XL.

\noindent\textbf{Visual Attention.}
The ablation study result in Table \ref{table:table2} shows that the visual attention mechanism is effective. It helps improve the performances of models, i.e., $+2.95$\% for GPT-2, $+0.78$\% for GPT-2-M, $+0.22$\% for GPT-2-L, and $+2$\% for GPT-2-XL.

\noindent\textbf{Interactive Attention.}
According to Table \ref{table:table2}, the proposed interactive attention network is effective and it improves the models' performances, i.e., $+1.35$\% for GPT-2, $+1.28$\% for GPT-2-M, $+0.5$\% for GPT-2-L, and $+2.49$\% for GPT-2-XL.

Since the above attentions help leverage the importance of features in a high dimensional space more effectively, it could help the model converge to a better local optima. Based on the above ablation study, it is concluded that all of the proposed attentions are effective.

\vspace{+10pt}\noindent\textbf{4.4 Effectiveness Analysis of Attentive Contextualized Word Representations}

\begin{figure*}[!ht]
  \begin{subfigure}[b]{1.0\textwidth}
    \includegraphics[width=0.98\linewidth]{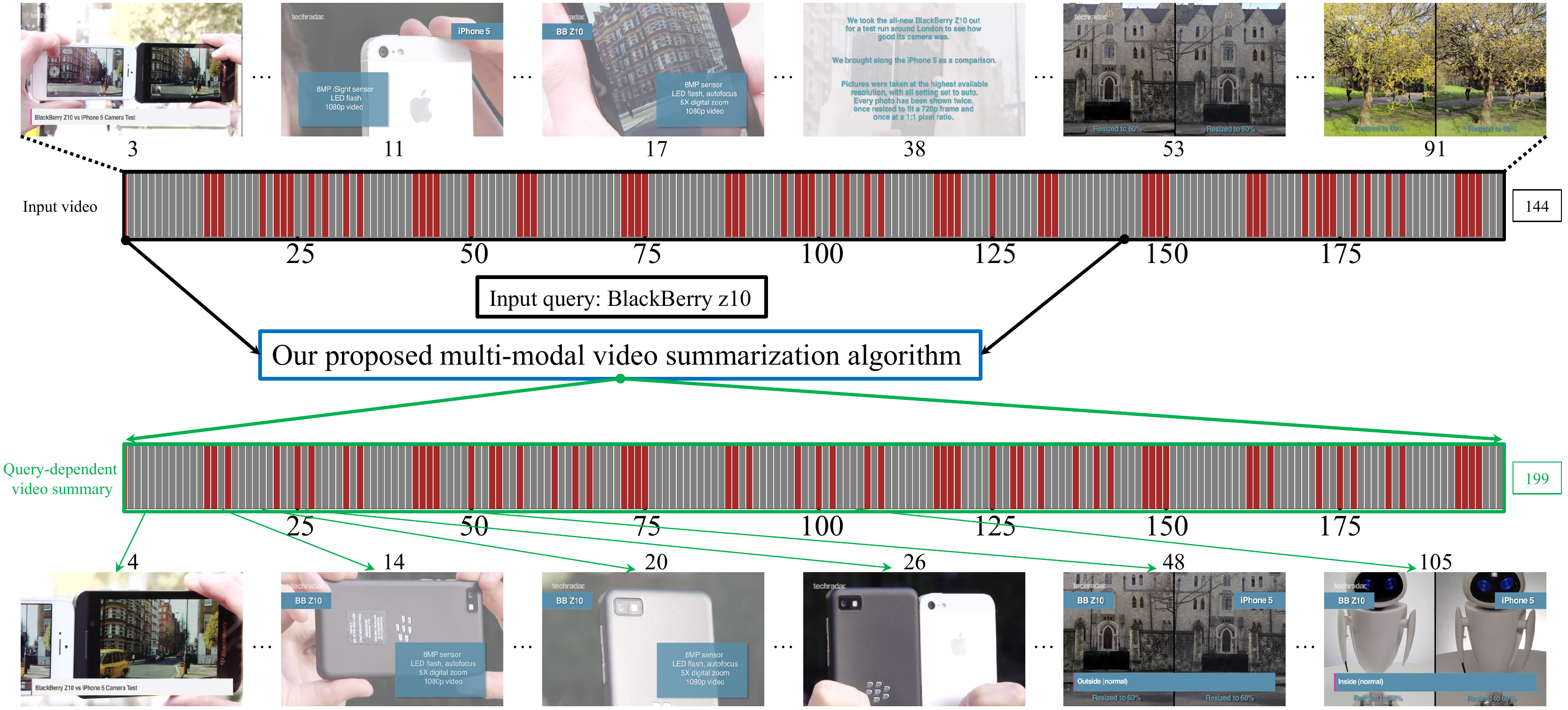}
    \caption{Input query: ``BlackBerry z10''. Correct number of score prediction $/$ Total number of frames = $144/199$.}
     \vspace{+0.5cm}
  \end{subfigure}
  \hfill
  \begin{subfigure}[b]{1.0\textwidth}
    \includegraphics[width=0.98\linewidth]{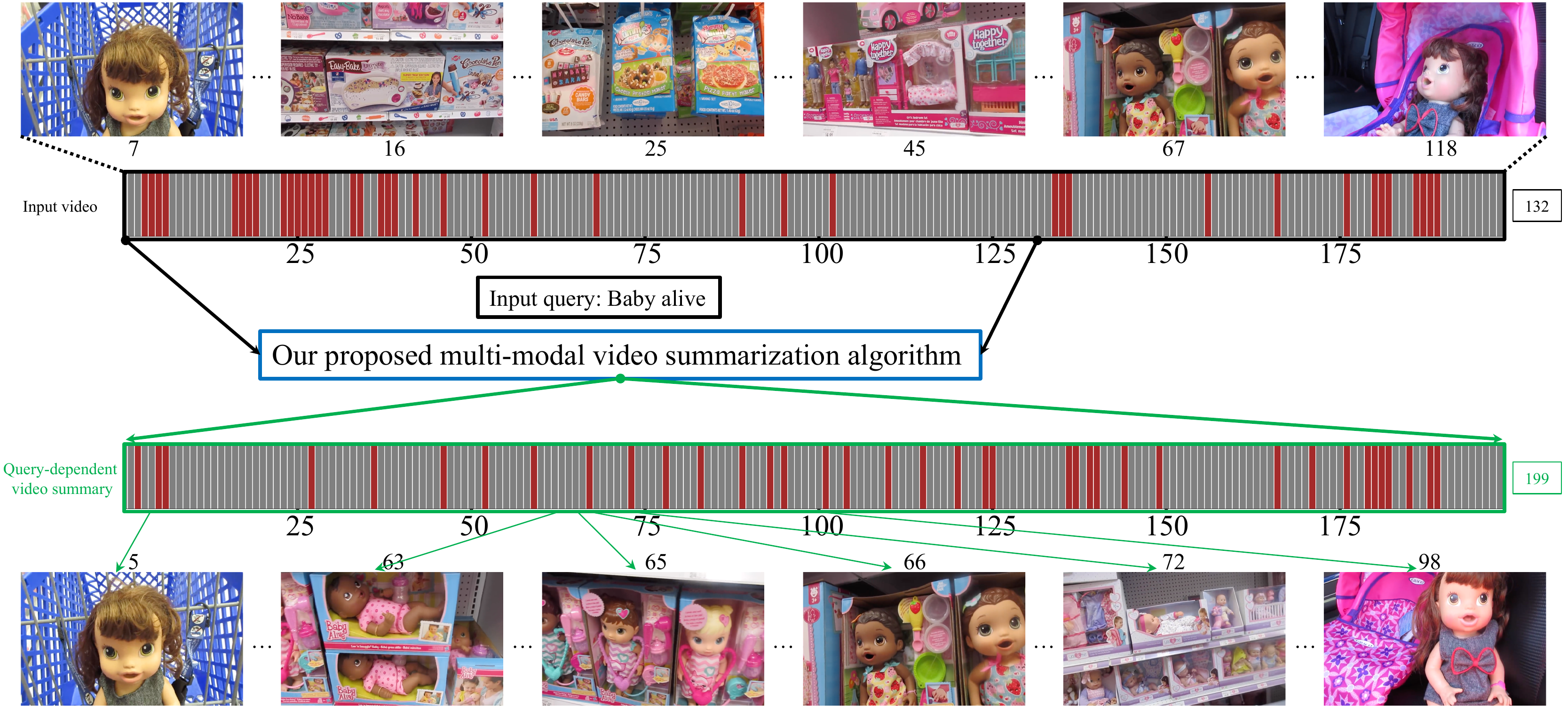}
    \caption{Input query: ``Baby alive''. Correct number of score prediction $/$ Total number of frames = $132/199$.}
  \end{subfigure}
\vspace{-0.7cm}
  \caption{Examples of the proposed multi-modal video summarization. Results are query-dependent video summaries (in green). The first two rows in example (a) represent the input video visualization and the corresponding ground truth frame-based score annotations, respectively. The last two rows in example (a) represent the prediction visualization of frame-based scores and the partial query-dependent video summary visualization, respectively. In each frame-based score pattern, gray denotes ``selected frames'' and red denotes ``not selected frames''. $144$ in example (a) denotes the video length before video preprocessing and $199$ denotes the video length after the video preprocessing. Refer to Subsection 4.2 for more details. Indices of the visualized selected frames are displayed in the figure. The same explanation from example (a) is applied to example (b).}
\label{fig:figure3}
\vspace{-0.1cm}
\end{figure*}

The authors of \cite{huang2020query} have proposed a state-of-the-art model, based on their newly introduced multi-modal video summarization benchmark \cite{huang2020query}. To show the effectiveness of the proposed attentive contextualized approach, the performance of the model is compared to \cite{huang2020query}. According to Table \ref{table:table3}, the result shows that the proposed method beats the state-of-the-art. The main reason is that the embedding of the multi-modal inputs, i.e., the text-based query and the video, is more effective than \cite{huang2020query}. This also shows that contextualized word representations are better than BoW used in \cite{huang2020query}. Qualitative results are demonstrated in Figure \ref{fig:figure3}.

\section{Conclusion}

In this work, a new end-to-end deep model is proposed in order to tackle the multi-modal video summarization problem. The proposed model consists of a contextualized video summary controller, a textual attention mechanism, a visual attention mechanism, an interactive attention network, and a video summary generator. Experimental results show that the proposed model is effective and achieves state-of-the-art performance, $+5.88$\% in terms of accuracy and $+4.06$\% in terms of F1-score. Since video data does not only have a visual channel but also a speech channel, modeling the speech input based on the transformer-encoder and transformer-decoder is an idea that could be explored in the future.

\begin{acks}
This project has received funding from the European Union’s Horizon 2020 research and innovation programme under the Marie Skłodowska-Curie grant agreement No 765140.
\end{acks}

%
\bibliographystyle{ACM-Reference-Format}
\bibliography{sample-base}

%
\appendix

\end{document}